\documentclass[11pt,letterpaper]{article}
\pdfoutput=1
\usepackage{emnlp2017}
\usepackage{times}
\usepackage{latexsym}

\emnlpfinalcopy

\def\emnlppaperid{1254}

\usepackage{amsmath}
\usepackage{multirow}
\usepackage{basecommon} 
\usepackage{array}
\newcolumntype{P}[1]{>{\centering\arraybackslash}p{#1}}
\newcolumntype{T}[1]{>{\raggedright\arraybackslash}p{#1}}

\title{Adapting Sequence Models for Sentence Correction}

\author{Allen Schmaltz{\textsuperscript{*}} \And Yoon Kim \And Alexander M. Rush \And Stuart M. Shieber \AND
Harvard University \\
{\tt \footnotesize \{schmaltz@fas,yoonkim@seas,srush@seas,shieber@seas\}.harvard.edu} \\
{\normalfont{\footnotesize\textsuperscript{*}Part of this work was completed while as an intern at Rakuten.}}
}

\date{}

\begin{document}

\maketitle

\begin{abstract}
In a controlled experiment of sequence-to-sequence approaches for the task of sentence correction, we find that character-based models are generally more effective than word-based models and models that encode subword information via convolutions, and that modeling the output data as a series of diffs improves effectiveness over standard approaches. Our strongest sequence-to-sequence model improves over our strongest phrase-based statistical machine translation model, with access to the same data, by 6 $M^2$ (0.5 GLEU) points. Additionally, in the data environment of the standard CoNLL-2014 setup, we demonstrate that modeling (and tuning against) diffs yields similar or better $M^2$ scores with simpler models and/or significantly less data than previous sequence-to-sequence approaches.
 \end{abstract}

\section{Introduction}
 The task of \emph{sentence correction} is to convert a natural language sentence that may or may not have errors into a corrected version. The task is envisioned as a component of a learning tool or writing-assistant, and has seen increased interest since 2011 driven by a  series of shared tasks  \cite{DaleAndKilgarriff-2011-HOOPilot,DaleEtAl-2012-HOO2012PrepAndDetErrors,NgEtAl-2013-CoNLL-SharedTask2013,NgEtAl-2014-SharedTask2014}. 

Most recent work on language correction has focused on the data provided by the CoNLL-2014 shared task \cite{NgEtAl-2014-SharedTask2014}, a set of corrected essays by second-language learners. The CoNLL-2014 data consists of only around 60,000 sentences, and as such, competitive systems have made use of large amounts of corrected text without annotations, and in some cases lower-quality crowd-annotated data, in addition to the shared data. In this data environment, it has been suggested that statistical phrase-based machine translation (MT) with task-specific features is the state-of-the-art for the task \cite{junczysdowmunt-grundkiewicz:2016:EMNLP2016}, outperforming word- and character-based sequence-to-sequence models \cite{yuan-briscoe:2016:N16-1,XieEtAl.2016-arxiv-NLCwithCharAttention,JiEtal2017arXiv-NestedAttention}, phrase-based systems with neural features \cite{ChollampattElAl-2016-SMTwithNNfeatures,chollampatt-hoang-ng:2016:EMNLP2016}, re-ranking output from phrase-based systems \cite{HoangEtAl2016-IJCAI-NbestSMTReranking}, and combining phrase-based systems with classifiers trained for hand-picked subsets of errors \cite{rozovskaya-roth:2016:P16-1}.

We revisit the comparison across translation approaches for the correction task in light of the Automated Evaluation of Scientific Writing (AESW) 2016 dataset, a correction dataset containing over 1 million sentences, holding constant the training data across approaches. The dataset was previously proposed for the distinct binary classification task of grammatical error identification. 

Experiments demonstrate that pure character-level sequence-to-sequence models are more effective on AESW than word-based models and models that encode subword information via convolutions over characters, and that representing the output data as a series of \emph{diffs} significantly increases effectiveness on this task. Our strongest character-level model achieves statistically significant improvements over our strongest phrase-based statistical machine translation model by 6 $M^2$ (0.5 GLEU) points, with additional gains when including domain information. Furthermore, in the partially crowd-sourced data environment of the standard CoNLL-2014 setup in which there are comparatively few professionally annotated sentences, we find that tuning against the tags marking the diffs yields similar or superior effectiveness relative to existing sequence-to-sequence approaches despite using significantly less data, with or without using secondary models. All code is available at \url{https://github.com/allenschmaltz/grammar}.

\section{Background and Methods}
\paragraph{Task} 
We follow recent work and treat the task of sentence correction as translation from a source sentence (the unedited sentence) into a target sentence (a corrected version in the same language as the source). We do not make a distinction between grammatical and stylistic corrections.

We assume a vocabulary $\mcV$ of natural language word types (some of which have orthographic errors).
Given a sentence $\mathbf{s} = [s_1 \cdots s_I]$, where $s_i \in \mcV$ is the $i$-th token of the sentence of length $I$, we seek to predict the corrected target sentence $\mathbf{t} = [t_1 \cdots t_J]$, where $t_j \in \mcV$ is the $j$-th token of the corrected sentence of length $J$.
We are given both $\mathbf{s}$ and $\mathbf{t}$ for supervised training in the standard setup. At test time, we are only given access to sequence $\mathbf{s}$. We learn to predict sequence $\mathbf{t}$ (which is often identical to $\mathbf{s}$).

\paragraph{Sequence-to-sequence}
 
We explore word and character variants of the sequence-to-sequence framework.
 We use a standard word-based model (\textsc{Word}), similar to that of \newcite{luong-pham-manning:2015:EMNLP}, as well as a model that uses a convolutional neural network (CNN) and a highway network over characters (\textsc{CharCNN}), based on the work of \newcite{KimEtAl-2016-CharLM}, instead of word embeddings as the input to the encoder and decoder. With both of these models, predictions are made at the word level. We also consider the use of bidirectional versions of these encoders (\textsc{+Bi}). 
 
 Our character-based model (\textsc{Char+Bi}) 
 follows the architecture of the \textsc{Word+Bi} model, but the input and output consist of characters rather than words. In this case, the input and output sequences are converted to a series of characters and whitespace delimiters. The output sequence is converted back to $\mathbf{t}$ prior to evaluation.
 
The \textsc{Word} models encode and decode over a closed vocabulary (of the 50k most frequent words); the \textsc{CharCNN} models encode over an open vocabulary and decode over a closed vocabulary; and the \textsc{Char} models encode and decode over an open vocabulary.  

Our contribution is to investigate the impact of sequence-to-sequence approaches (including those not considered in previous work) in a series of controlled experiments, holding the data constant. In doing so, we demonstrate that on a large, professionally annotated dataset, the most effective sequence-to-sequence approach can significantly outperform a state-of-the-art SMT system without augmenting the sequence-to-sequence model with a secondary model to handle low-frequency words \cite{yuan-briscoe:2016:N16-1} or an additional model to improve precision or intersecting a large language model \cite{XieEtAl.2016-arxiv-NLCwithCharAttention}. We also demonstrate improvements over these previous sequence-to-sequence approaches on the CoNLL-2014 data and competitive results with \newcite{JiEtal2017arXiv-NestedAttention}, despite using significantly less data. 

The work of \newcite{schmaltz-EtAl:2016:BEA11} applies \textsc{Word} and \textsc{CharCNN} models to the distinct binary classification task of error identification.

\paragraph{Additional Approaches}

The standard formulation of the correction task is to model the output sequence as $\mathbf{t}$ above. Here, we also propose modeling the diffs between $\mathbf{s}$ and $\mathbf{t}$. The diffs are provided in-line within $\mathbf{t}$ and are described via tags marking the starts and ends of insertions and deletions, with replacements represented as deletion-insertion pairs, as in the following example selected from the training set:  ``Some key points are worth~\textless del\textgreater~emphasiz~ \textless /del\textgreater ~\textless ins\textgreater~emphasizing~\textless /ins\textgreater~.''. Here, ``emphasiz'' is replaced with ``emphasizing''. The models, including the \textsc{Char} model, treat each tag as a single, atomic token. 

The diffs enable a means of tuning the model's propensity to generate corrections by modifying the probabilities generated by the decoder for the 4 diff tags, which we examine with the CoNLL data. We include four bias parameters associated with each diff tag, and run a grid search between 0 and 1.0 to set their values based on the tuning set.

It is possible for models with diffs to output invalid target sequences (for example, inserting a word without using a diff tag). To fix this, a deterministic post-processing step is performed (greedily from left to right) that returns to source any non-source tokens outside of insertion tags. Diffs are removed prior to evaluation. We indicate models that \textit{do not} incorporate target diff annotation tags with the designator \textsc{\textbf{--}diffs}.

The AESW dataset provides the paragraph context and a journal domain (a classification of the document into one of nine subject categories) for each sentence.\footnote{The paragraphs are shuffled for purposes of obfuscation, 
so document-level context is not available.} For the sequence-to-sequence models we propose modeling the input and output sequences with a special initial token representing the journal domain (\textsc{+dom}).\footnote{Characteristics of the dataset preclude experiments with additional paragraph context features. (See Appendix A.)} 

\begin{table}
\centering
\footnotesize
\begin{tabular}{lcccc}
\toprule
 & \multicolumn{2}{c}{GLEU} & \multicolumn{2}{c}{$M^2$} \\
Model & Dev & Test & Dev & Test \\
\midrule
No Change & 89.68 & 89.45 & 00.00 & 00.00 \\
\midrule
\textsc{SMT\textbf{--}diffs+$M^2$} & 90.44 &$-$ &38.55 & $-$ \\
\textsc{SMT\textbf{--}diffs+BLEU} & 90.90 & $-$ & 37.66 & $-$ \\
\textsc{Word+Bi\textbf{--}diffs} & 91.18 & $-$ & 38.88 & $-$ \\
\textsc{Char+Bi\textbf{--}diffs} & 91.28 & $-$ & 40.11 & $-$ \\
\midrule
\textsc{SMT+BLEU} & 90.95 & 90.70 & 38.99 & 38.31 \\
\textsc{Word+Bi} & 91.34 & 91.05 & 43.61 & 42.78\\
\textsc{CharCNN} & 91.23 & 90.96 & 42.02 & 41.21 \\
\textsc{Char+Bi} & \textbf{91.46} & \textbf{91.22} & \textbf{44.67} & \textbf{44.62} \\
\midrule
\textsc{Word+dom} & 91.25 & $-$ & 43.12 & $-$\\
\textsc{Word+Bi+dom} & 91.45 & $-$ &44.33 & $-$ \\
\textsc{CharCNN+Bi+dom} & 91.15 & $-$ & 40.79 & $-$\\
\textsc{CharCNN+dom} & 91.35 & $-$ & 43.94 & $-$\\
\textsc{Char+Bi+dom} & \textbf{91.64} & \textbf{91.39} & \textbf{47.25} & \textbf{46.72} \\
\bottomrule
\end{tabular}
\caption[Main results]{\small{AESW development/test set correction results. GLEU and $M^2$ differences on test are statistically significant via paired bootstrap resampling \cite{koehn:2004:EMNLP,graham-mathur-baldwin:2014:W14-33} at the 0.05 level, resampling the full set 50 times.}} 
\label{table:dev-results}
\end{table}

\section{Experiments}
\label{sec:experiments}

\paragraph{Data}
AESW \cite{Daudaravicius2016-dataset,daudaravicius-EtAl:2016:BEA11} consists of sentences taken from academic articles annotated with corrections by professional editors used for the AESW shared task. The training set contains 1,182,491 sentences, of which 460,901 sentences have edits. We set aside a 9,947 sentence sample from the original development set for tuning (of which 3,797 contain edits), and use the remaining 137,446 sentences as the dev set\footnote{The dev set contains 13,562 unique deletion types, 29,952 insertion types, and 39,930 replacement types.} (of which 53,502 contain edits). The test set contains 146,478 sentences.

The primary focus of the present study is conducting controlled experiments on the AESW dataset, but we also investigate results on the CoNLL-2014 shared task data in light of recent neural results \cite{JiEtal2017arXiv-NestedAttention} and to serve as a baseline of comparison against existing sequence-to-sequence approaches \cite{yuan-briscoe:2016:N16-1,XieEtAl.2016-arxiv-NLCwithCharAttention}. We use the common sets of public data appearing in past work for training: the National University of Singapore (NUS) Corpus of Learner English (NUCLE) \cite{dahlmeier-ng-wu:2013:BEA8} and the publicly available Lang-8 data \cite{tajiri-komachi-matsumoto:2012:ACL2012short,mizumoto-EtAl:2012:POSTERS}. The Lang-8 dataset of corrections is large\footnote{about 1.4 million sentences after filtering} but is crowd-sourced\footnote{derived from the Lang-8 language-learning website} and is thus of a different nature than the professionally annotated AESW and NUCLE datasets. We use the revised CoNLL-2013 test set as a tuning/dev set and the CoNLL-2014 test set (without alternatives) for testing. We do not make use of the non-public Cambridge Learner Corpus (CLC) \cite{Nicholls2003-CLC}, which contains over 1.5 million sentence pairs. 

\paragraph{Evaluation} We follow past work and use the Generalized Language Understanding Evaluation (GLEU) \cite{napoles2016gleu} and MaxMatch ($M^2$) metrics \cite{DahlmeierEtAl2012-M2}.

\paragraph{Parameters}
All our models, implemented with OpenNMT \cite{2017opennmt}, are $2$-layer LSTMs with $750$ hidden units. For the \textsc{Word} model,
the word embedding size is also set to $750$, while for the \textsc{CharCNN} and \textsc{Char} models
we use a character embedding size of $25$. The \textsc{CharCNN} model has a convolutional
layer with $1000$ filters of width $6$ followed by max-pooling, which is fed into a $2$-layer highway network. Additional training details are provided in Appendix A. For AESW, the \textsc{Word+Bi} model contains around 144 million parameters, the \textsc{CharCNN+Bi} model around 79 million parameters, and the \textsc{Char+Bi} model around 25 million parameters.

\paragraph{Statistical Machine Translation}
 As a baseline of comparison, we experiment with a phrase-based machine translation approach (\textsc{SMT}) shown to be state-of-the-art for the CoNLL-2014 shared task data in previous work \cite{junczysdowmunt-grundkiewicz:2016:EMNLP2016}, which adds task specific features and the $M^2$ metric as a scorer to the Moses statistical machine translation system. The \textsc{SMT} model follows the training, parameters, and dense and sparse task-specific features that generate state-of-the-art results for CoNLL-2014 shared task data, as implemented in publicly available code.\footnote{SRI International provided access to SRILM \cite{Stolcke02srilm} for running \newcite{junczysdowmunt-grundkiewicz:2016:EMNLP2016}} However, to compare models against the same training data, we remove language model features associated with external data.\footnote{We found that including the features and data associated with the large language models of \newcite{junczysdowmunt-grundkiewicz:2016:EMNLP2016}, created from Common Crawl text filtered against the NUCLE corpus, \textit{hurt} effectiveness for the phrase-based models. This is likely a reflection of the domain specific nature of the academic text and LaTeX holder symbols appearing in the text. Here, we conduct controlled experiments without introducing additional domain-specific monolingual data.} We experiment with tuning against $M^2$ (\textsc{+$M^2$}) and BLEU (\textsc{+BLEU}). Models trained with diffs were only tuned with BLEU, since the tuning pipeline from previous work is not designed to handle removing such annotation tags prior to $M^2$ scoring. 

\section{Results and Analysis: AESW}

\begin{table*}
\centering
\footnotesize
\begin{tabular}{lccccccc}
\toprule
 \multicolumn{7}{c}{Replacement Error Type (out of 39,930) -- Frequency relative to training}  \\
\midrule
Model & Punctuation  & Articles & Other \\
& & &  $>100$ & $[5,100]$ &  $[2,5)$ & $1$ & $0$  \\
\midrule
Raw frequency in dev & 11507 & 1691 & 6788 & 8974 & 2271 & 1620 & 7079 \\ 
Number of unique instances & 371 & 367 & 215 & 2918 & 1510 & 1242 & 5819 \\ 
\midrule
\textsc{SMT+BLEU} & 56.03 & 16.41 & 44.57 & 36.17 & 39.46 & 31.93 & 0.00  \\ 
\midrule
\textsc{Word+Bi} & 56.13 & 18.58 & 55.38 & 44.33 & 18.79 & 6.38 & 0.77 \\  
\textsc{Word+Bi+dom} & 56.87 & 19.16 & 59.02 & 44.57 & 19.70 & 4.42 & 2.01 \\ 
\midrule
\textsc{CharCNN+dom} & 55.64 & 13.37 & 57.34 & 41.83 & 28.99 & 16.74 & 7.09 \\ 
\midrule
\textsc{Char+Bi}  & 58.71 & 28.40 & 55.34 & 44.59 & 28.98 & 24.48 & 14.14 \\ 
\textsc{Char+Bi+dom} & 58.93 & 27.64 & 59.32 & 46.08 & 32.82 & 26.48 & 18.66 \\
\bottomrule
\end{tabular}
\caption{\small{Micro $F_{0.5}$ scores on replacement errors on the dev set. Errors are grouped by `Punctuation', `Article', and `Other'. `Other' errors are further broken down based on frequency buckets on the training set, with errors grouped by the frequency in which they occur in the training set.}}
\label{table:replacementsTrainingFreq}
\end{table*}

\begin{table}
\centering
\small
\begin{tabular}{p{22mm}P{10mm}P{10mm}P{18mm}}
\toprule
& Deletions & Insertions & Replacements \\
\midrule
\textsc{SMT+BLEU} & 46.56 & 31.48 & 42.21\\ 
\midrule
\textsc{Word+Bi} & 47.75 & 38.31  & 46.02 \\ 
\textsc{Word+Bi+dom} & 47.78 & 39.00 & 47.29 \\ 
\midrule
\textsc{CharCNN+dom} & 48.30 & 39.57 & 46.24 \\ 
\midrule
\textsc{Char+Bi} & 49.05 & 37.17 & 48.55 \\
\textsc{Char+Bi+dom} & 50.20 & 42.51 & 50.39 \\ 
\bottomrule
\end{tabular}
\caption{\small{Micro $F_{0.5}$ scores across error types}}
\label{table:deletionsInsertionsTrainingFreq}
\end{table}

Table 1 shows the full set of experimental results on the AESW development and test data. 

The \textsc{Char+Bi+dom} model is stronger than the \textsc{Word+Bi+dom} and \textsc{CharCNN+dom} models by 2.9 $M^2$ (0.2 GLEU) and 3.3 $M^2$ (0.3 GLEU), respectively. 
The sequence-to-sequence models were also more effective than the \textsc{SMT} models, as shown in Table \ref{table:dev-results}. We find that training with target diffs is beneficial across all models, with an increase of about 5 $M^2$ points for the \textsc{Word+bi} model, for example. Adding \textsc{+dom} information slightly improves effectiveness across models.

We analyzed deletion, insertion, and replacement error types.
Table \ref{table:replacementsTrainingFreq} compares effectiveness across replacement errors. We found the \textsc{CharCNN+Bi} models were less effective than \textsc{CharCNN} variants in terms of GLEU and $M^2$, and the strongest \textsc{CharCNN} models were eclipsed by the \textsc{Word+Bi} models in terms of the GLEU and $M^2$ scores. However, Table \ref{table:replacementsTrainingFreq} shows \textsc{CharCNN+dom} is stronger on lower frequency replacements than \textsc{Word} models. The \textsc{Char+Bi+dom} model is relatively strong on article and punctuation replacements, as well as errors appearing with low frequency in the training set and overall across deletion and insertion error types, which are summarized in Table \ref{table:deletionsInsertionsTrainingFreq}. 

\paragraph{Errors never occurring in training} 
The comparatively high Micro $F_{0.5}$ score (18.66) for the \textsc{Char+Bi+dom} model on replacement errors (Table  \ref{table:replacementsTrainingFreq}) never occurring in training is a result of a high precision (92.65) coupled with a low recall (4.45). This suggests some limited capacity to generalize to items not seen in training. A selectively chosen example is the replacement from ``discontinous'' to ``discontinuous'', which never occurs in training. However, similar errors of low edit distance also occur once in the dev set and never in training, but the \textsc{Char+Bi+dom} model never correctly recovers many of these errors, and many of the correctly recovered errors are minor changes in capitalization or hyphenation.

\paragraph{Error frequency} About 39\% of the AESW training sentences have errors, and of those sentences, on average, 2.4 words are involved in changes in deletions, insertions, or replacements (i.e., the count of words occurring between diff tags) per sentence. In the NUCLE data, about 37\% of the sentences have errors, of which on average, 5.3 words are involved in changes. On the AESW dev set, if we only consider the 9545 sentences in which 4 or more words are involved in a change (average of 5.8 words in changes per sentence), the \textsc{Char+Bi} model is still more effective than \textsc{SMT+BLEU}, with a GLEU score of 67.21 vs. 65.34. The baseline GLEU score (No Change) is 60.86, reflecting the greater number of changes relative to the full dataset (cf. Table 1).

\paragraph{Re-annotation}
The AESW dataset only provides 1 annotation for each sentence, so we perform a small re-annotation of the data to gauge effectiveness in the presence of multiple annotations. We collected 3 outputs (source, gold, and generated sentences from the \textsc{Char+Bi+dom} model) for 200 randomly sampled sentences, re-annotating to create 3 new references for each sentence. The GLEU scores for the 200 original source, \textsc{Char+Bi+dom}, and original gold sentences evaluated against the 3 new references were 79.79, 81.72, and 84.78, respectively, suggesting that there is still progress to be made on the task relative to human levels of annotation. 

\section{Results and Analysis: CoNLL}

\begin{table}
\centering
\small
\begin{tabular}{lccc}
\toprule
& Precision & Recall & $F_{0.5}$ \\
\midrule
\textsc{Word+Bi\textbf{--}diffs} & 65.36 & 6.19 & 22.45 \\
\textsc{Word+Bi}, before tuning & 72.34 & 0.97 & 4.60 \\ 
\textsc{Word+Bi}, after tuning & 46.66 & 15.35 & 33.14 \\ 
\bottomrule
\end{tabular}
\caption{\small{$M^2$ scores on the CoNLL-2013 set.}}
\label{table:conll-tuning}
\end{table}

Table \ref{table:conll-tuning} shows the results on the CoNLL dev set, and Table \ref{table:conll-data-and-test} contains the final test results.

Since the CoNLL data does not contain enough data for training neural models, previous works add the crowd-sourced Lang-8 data; however, this data is not professionally annotated. Since the distribution of corrections differs between the dev/test and training sets, we need to tune the precision and recall.

As shown in Table \ref{table:conll-tuning}, \textsc{Word+Bi} effectiveness increases significantly by tuning the weights\footnote{In contrast, in early experiments on AESW, tuning yielded negligible improvements.} assigned to the diff tags on the CoNLL-2013 set\footnote{The single model with highest $M^2$ score was then run on the test set. Here, a single set is used for tuning and dev.}. Note that we are tuning the weights on this same CoNLL-2013 set. Without tuning, the model very rarely generates a change, albeit with a high precision. After tuning, it exceeds the effectiveness of \textsc{Word+Bi\textbf{--}diffs}. The comparatively low effectiveness of \textsc{Word+Bi\textbf{--}diffs} is consistent with past sequence-to-sequence approaches utilizing data augmentation, additional annotated data, and/or secondary models to achieve competitive levels of effectiveness.

Table \ref{table:conll-data-and-test} shows that \textsc{Word+Bi} is within 0.2 $M^2$ of \newcite{JiEtal2017arXiv-NestedAttention}, despite using over 1 million fewer sentence pairs, and exceeds the $M^2$ scores of \newcite{XieEtAl.2016-arxiv-NLCwithCharAttention} and \newcite{yuan-briscoe:2016:N16-1} without the secondary models of those systems. We hypothesize that further gains are possible utilizing the CLC data and moving to the character model. (The character model is omitted here due to the long training time of about 4 weeks.) Notably, SMT systems (with LMs) are still more effective than reported sequence-to-sequence results, as in \newcite{JiEtal2017arXiv-NestedAttention}, on CoNLL.\footnote{For reference, the reported $M^2$ results of the carefully optimized SMT system of  \newcite{junczysdowmunt-grundkiewicz:2016:EMNLP2016} trained on NUCLE and Lang-8, with parameter vectors averaged over multiple runs, with a Wikipedia LM is 45.95 and adding a Common Crawl LM is 49.49. We leave to future work the intersection of a LM for the CoNLL environment and more generally, whether these patterns hold in the presence of additional monolingual data.}
 
\begin{table}
\centering
\small
\begin{tabular}{lT{26mm}c}
\toprule
& Data & $M^2$ \\
\midrule
\newcite{yuan-briscoe:2016:N16-1} & CLC$^*$ & 39.90 \\
\midrule
\newcite{XieEtAl.2016-arxiv-NLCwithCharAttention} & NUCLE, Lang-8, Common Crawl LM & 40.56 \\
\midrule
\newcite{JiEtal2017arXiv-NestedAttention} & NUCLE, Lang-8, CLC$^*$ & 41.53 \\
\midrule
\textsc{Word+Bi\textbf{--}diffs} & NUCLE, Lang-8 & 35.73 \\
\textsc{Word+Bi} & NUCLE, Lang-8 & 41.37  \\ 
\bottomrule
\end{tabular}
\caption{\small{$M^2$ scores on the CoNLL-2014 test set and data used for recent sequence-to-sequence based systems. Results for previous works are those reported by the original authors. $^*$CLC is proprietary.}
\label{table:conll-data-and-test}}
\end{table}

\section{Conclusion}

Our experiments demonstrate that on a large, professionally annotated dataset, a sequence-to-sequence character-based model of diffs can lead to considerable effectiveness gains over a state-of-the-art SMT system with task-specific features, ceteris paribus. Furthermore, in the crowd-sourced environment of the CoNLL data, in which there are comparatively few professionally annotated sentences in training, modeling diffs enables a means of tuning that improves the effectiveness of sequence-to-sequence models for the task.

\bibliography{emnlp2017_sentence_correction_with_appendix}

\begin{thebibliography}{28}
\expandafter\ifx\csname natexlab\endcsname\relax\def\natexlab#1{#1}\fi

\bibitem[{Chollampatt et~al.(2016{\natexlab{a}})Chollampatt, Hoang, and
  Ng}]{chollampatt-hoang-ng:2016:EMNLP2016}
Shamil Chollampatt, Duc~Tam Hoang, and Hwee~Tou Ng. 2016{\natexlab{a}}.
\newblock \href {https://aclweb.org/anthology/D16-1195} {Adapting grammatical
  error correction based on the native language of writers with neural network
  joint models}.
\newblock In \emph{Proceedings of the 2016 Conference on Empirical Methods in
  Natural Language Processing}, pages 1901--1911, Austin, Texas. Association
  for Computational Linguistics.

\bibitem[{Chollampatt et~al.(2016{\natexlab{b}})Chollampatt, Taghipour, and
  Ng}]{ChollampattElAl-2016-SMTwithNNfeatures}
Shamil Chollampatt, Kaveh Taghipour, and Hwee~Tou Ng. 2016{\natexlab{b}}.
\newblock \href {http://dl.acm.org/citation.cfm?id=3060832.3061008} {Neural
  network translation models for grammatical error correction}.
\newblock In \emph{Proceedings of the Twenty-Fifth International Joint
  Conference on Artificial Intelligence}, IJCAI'16, pages 2768--2774. AAAI
  Press.

\bibitem[{Dahlmeier and Ng(2012)}]{DahlmeierEtAl2012-M2}
Daniel Dahlmeier and Hwee~Tou Ng. 2012.
\newblock \href {http://dl.acm.org/citation.cfm?id=2382029.2382118} {Better
  evaluation for grammatical error correction}.
\newblock In \emph{Proceedings of the 2012 Conference of the North American
  Chapter of the Association for Computational Linguistics: Human Language
  Technologies}, NAACL HLT '12, pages 568--572, Stroudsburg, PA, USA.
  Association for Computational Linguistics.

\bibitem[{Dahlmeier et~al.(2013)Dahlmeier, Ng, and
  Wu}]{dahlmeier-ng-wu:2013:BEA8}
Daniel Dahlmeier, Hwee~Tou Ng, and Siew~Mei Wu. 2013.
\newblock \href {http://www.aclweb.org/anthology/W13-1703} {Building a large
  annotated corpus of learner english: The nus corpus of learner english}.
\newblock In \emph{Proceedings of the Eighth Workshop on Innovative Use of NLP
  for Building Educational Applications}, pages 22--31, Atlanta, Georgia.
  Association for Computational Linguistics.

\bibitem[{Dale et~al.(2012)Dale, Anisimoff, and
  Narroway}]{DaleEtAl-2012-HOO2012PrepAndDetErrors}
Robert Dale, Ilya Anisimoff, and George Narroway. 2012.
\newblock \href {http://dl.acm.org/citation.cfm?id=2390384.2390390} {Hoo 2012:
  A report on the preposition and determiner error correction shared task}.
\newblock In \emph{Proceedings of the Seventh Workshop on Building Educational
  Applications Using NLP}, pages 54--62, Stroudsburg, PA, USA. Association for
  Computational Linguistics.

\bibitem[{Dale and Kilgarriff(2011)}]{DaleAndKilgarriff-2011-HOOPilot}
Robert Dale and Adam Kilgarriff. 2011.
\newblock \href {http://dl.acm.org/citation.cfm?id=2187681.2187725} {Helping
  our own: The hoo 2011 pilot shared task}.
\newblock In \emph{Proceedings of the 13th European Workshop on Natural
  Language Generation}, ENLG '11, pages 242--249, Stroudsburg, PA, USA.
  Association for Computational Linguistics.

\bibitem[{Daudaravicius(2016)}]{Daudaravicius2016-dataset}
Vidas Daudaravicius. 2016.
\newblock Automated evaluation of scientific writing data set (version 1.2)
  [data file].
\newblock VTeX.

\bibitem[{Daudaravicius et~al.(2016)Daudaravicius, Banchs, Volodina, and
  Napoles}]{daudaravicius-EtAl:2016:BEA11}
Vidas Daudaravicius, Rafael~E. Banchs, Elena Volodina, and Courtney Napoles.
  2016.
\newblock \href {http://www.aclweb.org/anthology/W16-0506} {A report on the
  automatic evaluation of scientific writing shared task}.
\newblock In \emph{Proceedings of the 11th Workshop on Innovative Use of NLP
  for Building Educational Applications}, pages 53--62, San Diego, CA.
  Association for Computational Linguistics.

\bibitem[{Durrani et~al.(2013)Durrani, Fraser, Schmid, Hoang, and
  Koehn}]{durrani-EtAl:2013:Short}
Nadir Durrani, Alexander Fraser, Helmut Schmid, Hieu Hoang, and Philipp Koehn.
  2013.
\newblock \href {http://www.aclweb.org/anthology/P13-2071} {Can markov models
  over minimal translation units help phrase-based smt?}
\newblock In \emph{Proceedings of the 51st Annual Meeting of the Association
  for Computational Linguistics (Volume 2: Short Papers)}, pages 399--405,
  Sofia, Bulgaria. Association for Computational Linguistics.

\bibitem[{Graham et~al.(2014)Graham, Mathur, and
  Baldwin}]{graham-mathur-baldwin:2014:W14-33}
Yvette Graham, Nitika Mathur, and Timothy Baldwin. 2014.
\newblock \href {http://aclweb.org/anthology/W/W14/W14-3333.pdf} {Randomized
  significance tests in machine translation}.
\newblock In \emph{Proceedings of the Ninth Workshop on Statistical Machine
  Translation}, pages 266--274, Baltimore, Maryland, USA. Association for
  Computational Linguistics.

\bibitem[{Hoang et~al.(2016)Hoang, Chollampatt, and
  Ng}]{HoangEtAl2016-IJCAI-NbestSMTReranking}
Duc~Tam Hoang, Shamil Chollampatt, and Hwee~Tou Ng. 2016.
\newblock \href {http://dl.acm.org/citation.cfm?id=3060832.3061013} {Exploiting
  n-best hypotheses to improve an smt approach to grammatical error
  correction}.
\newblock In \emph{Proceedings of the Twenty-Fifth International Joint
  Conference on Artificial Intelligence}, IJCAI'16, pages 2803--2809. AAAI
  Press.

\bibitem[{{Ji} et~al.(2017){Ji}, {Wang}, {Toutanova}, {Gong}, {Truong}, and
  {Gao}}]{JiEtal2017arXiv-NestedAttention}
J.~{Ji}, Q.~{Wang}, K.~{Toutanova}, Y.~{Gong}, S.~{Truong}, and J.~{Gao}. 2017.
\newblock \href {http://arxiv.org/abs/1707.02026} {{A Nested Attention Neural
  Hybrid Model for Grammatical Error Correction}}.
\newblock \emph{ArXiv e-prints}.

\bibitem[{Junczys-Dowmunt and
  Grundkiewicz(2016)}]{junczysdowmunt-grundkiewicz:2016:EMNLP2016}
Marcin Junczys-Dowmunt and Roman Grundkiewicz. 2016.
\newblock \href {https://aclweb.org/anthology/D16-1161} {Phrase-based machine
  translation is state-of-the-art for automatic grammatical error correction}.
\newblock In \emph{Proceedings of the 2016 Conference on Empirical Methods in
  Natural Language Processing}, pages 1546--1556, Austin, Texas. Association
  for Computational Linguistics.

\bibitem[{Kim et~al.(2016)Kim, Jernite, Sontag, and Rush}]{KimEtAl-2016-CharLM}
Yoon Kim, Yacine Jernite, David Sontag, and Alexander~M. Rush. 2016.
\newblock {C}haracter-{A}ware {N}eural {L}anguage {M}odels.
\newblock In \emph{Proceedings of AAAI}.

\bibitem[{{Klein} et~al.(){Klein}, {Kim}, {Deng}, {Senellart}, and
  {Rush}}]{2017opennmt}
G.~{Klein}, Y.~{Kim}, Y.~{Deng}, J.~{Senellart}, and A.~M. {Rush}.
\newblock \href {http://arxiv.org/abs/1701.02810} {{OpenNMT: Open-Source
  Toolkit for Neural Machine Translation}}.
\newblock \emph{ArXiv e-prints}.

\bibitem[{Koehn(2004)}]{koehn:2004:EMNLP}
Philipp Koehn. 2004.
\newblock Statistical significance tests for machine translation evaluation.
\newblock In \emph{Proceedings of EMNLP 2004}, pages 388--395, Barcelona,
  Spain. Association for Computational Linguistics.

\bibitem[{Luong et~al.(2015)Luong, Pham, and
  Manning}]{luong-pham-manning:2015:EMNLP}
Minh-Thang Luong, Hieu Pham, and Christopher~D. Manning. 2015.
\newblock \href {http://aclweb.org/anthology/D15-1166} {Effective approaches to
  attention-based neural machine translation}.
\newblock In \emph{Empirical Methods in Natural Language Processing (EMNLP)},
  pages 1412--1421, Lisbon, Portugal. Association for Computational
  Linguistics.

\bibitem[{Mizumoto et~al.(2012)Mizumoto, Hayashibe, Komachi, Nagata, and
  Matsumoto}]{mizumoto-EtAl:2012:POSTERS}
Tomoya Mizumoto, Yuta Hayashibe, Mamoru Komachi, Masaaki Nagata, and Yuji
  Matsumoto. 2012.
\newblock \href {http://www.aclweb.org/anthology/C12-2084} {The effect of
  learner corpus size in grammatical error correction of {ESL} writings}.
\newblock In \emph{Proceedings of COLING 2012: Posters}, pages 863--872,
  Mumbai, India. The COLING 2012 Organizing Committee.

\bibitem[{Napoles et~al.(2016)Napoles, Sakaguchi, Post, and
  Tetreault}]{napoles2016gleu}
Courtney Napoles, Keisuke Sakaguchi, Matt Post, and Joel Tetreault. 2016.
\newblock \href {http://arxiv.org/abs/1605.02592} {{GLEU} without tuning}.
\newblock \emph{eprint arXiv:1605.02592 [cs.CL]}.

\bibitem[{Ng et~al.(2014)Ng, Wu, Briscoe, Hadiwinoto, Susanto, and
  Bryant}]{NgEtAl-2014-SharedTask2014}
Hwee~Tou Ng, Siew~Mei Wu, Ted Briscoe, Christian Hadiwinoto, Raymond~Hendy
  Susanto, and Christopher Bryant. 2014.
\newblock \href {http://www.aclweb.org/anthology/W/W14/W14-1701} {The
  {CoNLL}-2014 shared task on grammatical error correction}.
\newblock In \emph{Proceedings of the Eighteenth Conference on Computational
  Natural Language Learning: Shared Task}, pages 1--14, Baltimore, Maryland.
  Association for Computational Linguistics.

\bibitem[{Ng et~al.(2013)Ng, Wu, Wu, Hadiwinoto, and
  Tetreault}]{NgEtAl-2013-CoNLL-SharedTask2013}
Hwee~Tou Ng, Siew~Mei Wu, Yuanbin Wu, Christian Hadiwinoto, and Joel Tetreault.
  2013.
\newblock \href {http://www.aclweb.org/anthology/W13-3601} {The {CoNLL}-2013
  shared task on grammatical error correction}.
\newblock In \emph{Proceedings of the Seventeenth Conference on Computational
  Natural Language Learning: Shared Task}, pages 1--12, Sofia, Bulgaria.
  Association for Computational Linguistics.

\bibitem[{Nicholls(2003)}]{Nicholls2003-CLC}
Diane Nicholls. 2003.
\newblock The cambridge learner corpus: Error coding and analysis for
  lexicography and {ELT}.
\newblock In \emph{Proceedings of the Corpus Linguistics 2003 conference},
  pages 572--581.

\bibitem[{Rozovskaya and Roth(2016)}]{rozovskaya-roth:2016:P16-1}
Alla Rozovskaya and Dan Roth. 2016.
\newblock \href {http://www.aclweb.org/anthology/P16-1208} {Grammatical error
  correction: Machine translation and classifiers}.
\newblock In \emph{Proceedings of the 54th Annual Meeting of the Association
  for Computational Linguistics (Volume 1: Long Papers)}, pages 2205--2215,
  Berlin, Germany. Association for Computational Linguistics.

\bibitem[{Schmaltz et~al.(2016)Schmaltz, Kim, Rush, and
  Shieber}]{schmaltz-EtAl:2016:BEA11}
Allen Schmaltz, Yoon Kim, Alexander~M. Rush, and Stuart Shieber. 2016.
\newblock \href {http://www.aclweb.org/anthology/W16-0528} {Sentence-level
  grammatical error identification as sequence-to-sequence correction}.
\newblock In \emph{Proceedings of the 11th Workshop on Innovative Use of NLP
  for Building Educational Applications}, pages 242--251, San Diego, CA.
  Association for Computational Linguistics.

\bibitem[{Stolcke(2002)}]{Stolcke02srilm}
Andreas Stolcke. 2002.
\newblock Srilm -- an extensible language modeling toolkit.
\newblock In \emph{Proc. Intl. Conf. on Spoken Language Processing}, volume~2,
  pages 901--904, Denver.

\bibitem[{Tajiri et~al.(2012)Tajiri, Komachi, and
  Matsumoto}]{tajiri-komachi-matsumoto:2012:ACL2012short}
Toshikazu Tajiri, Mamoru Komachi, and Yuji Matsumoto. 2012.
\newblock \href {http://www.aclweb.org/anthology/P12-2039} {Tense and aspect
  error correction for esl learners using global context}.
\newblock In \emph{Proceedings of the 50th Annual Meeting of the Association
  for Computational Linguistics (Volume 2: Short Papers)}, pages 198--202, Jeju
  Island, Korea. Association for Computational Linguistics.

\bibitem[{Xie et~al.(2016)Xie, Avati, Arivazhagan, Jurafsky, and
  Ng}]{XieEtAl.2016-arxiv-NLCwithCharAttention}
Ziang Xie, Anand Avati, Naveen Arivazhagan, Dan Jurafsky, and Andrew~Y. Ng.
  2016.
\newblock \href {http://arxiv.org/abs/1603.09727} {Neural language correction
  with character-based attention}.
\newblock \emph{CoRR}, abs/1603.09727.

\bibitem[{Yuan and Briscoe(2016)}]{yuan-briscoe:2016:N16-1}
Zheng Yuan and Ted Briscoe. 2016.
\newblock \href {http://www.aclweb.org/anthology/N16-1042} {Grammatical error
  correction using neural machine translation}.
\newblock In \emph{Proceedings of the 2016 Conference of the North American
  Chapter of the Association for Computational Linguistics: Human Language
  Technologies}, pages 380--386, San Diego, California. Association for
  Computational Linguistics.

\end{thebibliography}
\bibliographystyle{emnlp_natbib}

\appendix

\section{Supplemental Material}
\label{sec:supplemental}

\paragraph{Additional Model Training and Inference Details} We provide additional replication details for our experiments here. Our code and related materials are available at the following url: \url{https://github.com/allenschmaltz/grammar}.

The training and tuning sizes of the AESW dataset are those after dropping sentences exceeding 126 tokens on the source or target side (in source sequences or target sequences with diff annotation tags) from the raw AESW dataset. All evaluation metrics on the development and test set are on the data without filtering based on sentence lengths.

As part of preprocessing, the sentences from the AESW XML are converted to Penn Treebank-style tokenization. Case is maintained and digits are not replaced with holder symbols for the sequence-to-sequence models. For the \textsc{SMT} models, the truecasing\footnote{Here, the truecase language model is created from the training $\mathbf{t}$ sequences (or where applicable, the target with diffs).} and tokenization pipeline of the publicly available code is used. For consistency, all model output and all reference files are converted to cased Moses-style tokenization prior to evaluation. 

For the \textsc{Char} model, the $L_2$-normalized gradients were constrained to be $\le 1$ (instead of $\le 5$ with the other models), and our learning rate schedule started the learning rate at 0.5 (instead of 1 for the other models) for stable training. The maximum sequence length of 421 was used for models given character sequences, which was equivalent to the maximum sequence length of 126 used for models given word sequences. The maximum sequence lengths were increased by 1 for the models with the \textsc{+dom} features. The training and tuning set sizes cited in Section 3 are the number of sentences from the raw dataset after dropping sentences exceeding these maximum sequence lengths.

In practice, we were able to train each of the purely character-based models (e.g., the \textsc{Char+Bi+dom} model) with a single NVIDIA Quadro P6000 GPU with 24 GB of memory in about 3 weeks with a batch size of 12.

For the sequence-to-sequence models, the closed vocabularies were restricted to the 50,000 most common tokens, and a single special $\mathsf{ \textless unk\textgreater}$ token was used for all remaining low frequency tokens. An $\mathsf{ \textless unk\textgreater}$ token generated in the target sentence by the \textsc{Word} and \textsc{CharCNN} models was replaced with the source token associated with the maximum attention weight. The ``open'' vocabularies were only limited to the space of characters seen in training.

For the phrase-based machine translation baseline model from the work of \newcite{junczysdowmunt-grundkiewicz:2016:EMNLP2016}, for dense features, we used the stateless edit distance features and the stateful Operation Sequence Model (OSM) of \newcite{durrani-EtAl:2013:Short}\footnote{The OSM features use the SRI Language Modeling Toolkit (SRILM) \cite{Stolcke02srilm}. 

}. Since for our controlled data experiments we removed the language model features associated with external data, we did not use the word-class language model feature, so for the sparse features, we used the set of edit operations on ``words with left/right context of maximum length 1 on words'' (set ``E0C10'' from the original paper), instead of those dependent on word classes.

The training and tuning splits for the phrase-based machine translation models were the same as for the sequence-to-sequence models. For tuning, we used Batch-Mira, setting the background corpus decay rate to 0.001, as in previous work. As in previous work, we repeated the tuning process multiple times (in this case, 5 times) and averaged the final weight vectors.

The sequence-to-sequence models were decoded with a beam size of 10. 

Decoding of the \textsc{SMT} models used the same approach of \newcite{junczysdowmunt-grundkiewicz:2016:EMNLP2016} (i.e., the open-source Moses decoder run with the cube pruning search algorithm).

In our experiments, we do not include additional paragraph context features, since the underlying AESW data appears to have been collected such that nearly all paragraphs (including those containing a single sentence) contain at least one error; thus, modeling paragraph information provides additional signal that seems unlikely to reflect real-world environments.

\paragraph{CoNLL-2014 Shared Task}
For training, we used the copy of the Lang-8 corpus distributed in the repo for the code of \newcite{junczysdowmunt-grundkiewicz:2016:EMNLP2016}: \url{https://github.com/grammatical/baselines-emnlp2016}. We filtered the Lang-8 data to remove duplicates and target sentences containing emoticon text, informal colloquial words (e.g., ``haha'', ``lol'', ``yay''), and non-ascii characters. Target sentences not starting with a capital letter were dropped, as were target sentences not ending in a period, question mark, exclamation mark, or quotation mark. (Target sentences ending in a parenthesis were dropped as they often indicate informal additional comments from the editor.) In the combined NUCLE and Lang-8 training set, source sentences longer than 79 tokens and target sentences longer than 100 tokens were dropped. This resulted in a training set with 1,470,992 sentences. Diffs were created using the Python class difflib.SequenceMatcher.

For tuning on the dev set\footnote{Previous work, such as \newcite{junczysdowmunt-grundkiewicz:2016:EMNLP2016}, also used the CoNLL-2013 set for tuning.}, a coarse grid search between 0 and 1.0 was used to set the four bias parameters associated with each diff tag. (Training was performed without re-weighting.) The bias parameter (in this case 0.7) yielding the highest $M^2$ score on the decoded dev set was chosen for use in evaluation of the final test set. The $M^2$ scores across the tuning runs on the dev set for the \textsc{Word+Bi} model are shown in Table \ref{table:conll-tuning-weights}.

\begin{table}
\centering
\small
\begin{tabular}{cccc}
\toprule
Bias parameter & Precision & Recall & $F_{0.5}$ \\
\midrule
0.0 & 72.34 & 0.97 & 4.60 \\ 
0.1 &  69.74 & 1.51 & 6.96 \\  
0.2 &  72.00 & 2.57 & 11.23 \\  
0.3 & 69.05 & 4.14 & 16.68 \\  
0.4 & 67.19 & 6.08 & 22.31 \\  
0.5 & 61.03 & 8.76 & 27.82 \\  
0.6 & 51.75 & 11.41 & 30.31 \\  
\textbf{0.7} & \textbf{46.66} & \textbf{15.35} & \textbf{33.14} \\  
0.8 & 40.01 & 18.68 & 32.57 \\  
0.9 &  34.49 & 22.08 & 31.00 \\  
1.0 & 30.17 & 24.90 & 28.94  \\  
\bottomrule
\end{tabular}
\caption{\small{$M^2$ scores on the CoNLL-2013 dev set for the \textsc{Word+Bi} model.}}
\label{table:conll-tuning-weights}
\end{table}

For future comparisons to our work on the CoNLL-2014 shared task data, we recommend using the preprocessing scripts provided in our code repo (\url{https://github.com/allenschmaltz/grammar}).

\paragraph{Table 2} The seven columns of Table 2 appearing in the main text are Micro $F_{0.5}$ scores for the errors within each frequency grouping. There are a total of 39,916 replacement changes. The replacements are grouped in regard to the changes within the opening and closing deletion tags and subsequent opening and closing insertion tags, as follows: (1) whether the replacement involves (on the deletion and/or insertion side) a single punctuation symbol (comma, colon, period, hyphen, apostrophe, quotation mark, semicolon, exclamation, question mark); (2) whether the replacement involves (on the deletion and/or insertion side) a single article (a, an, the); (3) non-article, non-punctuation grouped errors with frequency greater than 100 in the gold training data; (4) non-article, non-punctuation grouped errors with frequency less than or equal to 100 and greater than or equal to 5; (5) non-article, non-punctuation grouped errors with frequency less than 5 and greater than or equal to 2; (6) non-article, non-punctuation grouped errors with frequency equal to 1; (7) non-article, non-punctuation grouped errors that never occurred in the training data. Note that the large number of unique instances occurring for the ``punctuation'' and ``articles'' classes are a result of the large number of errors that can occur on the non-article, non-punctuation side of the replacement. The Micro $F_{0.5}$ scores are calculated by treating each individual error (rather than the agglomerated classes here) as binary classifications.

\end{document}